\documentclass[11pt]{article}

\usepackage[final]{acl}

\usepackage{times}
\usepackage{latexsym}
\usepackage{amsmath}
\usepackage[utf8]{inputenc}
\usepackage[T1]{fontenc}
\usepackage{textgreek}
\usepackage{url}
\usepackage{microtype}
\usepackage{inconsolata}
\usepackage{graphicx}
\usepackage{booktabs}
\usepackage{xcolor}
\usepackage[most]{tcolorbox}
\tcbuselibrary{listings,breakable}
\usepackage{hyperref}

\usepackage{booktabs}
\lstdefinestyle{promptstyle}{
  basicstyle=\ttfamily\footnotesize,
  breaklines=true,
  breakatwhitespace=false,
  columns=fullflexible,
  keepspaces=true,
  showstringspaces=false,
  tabsize=2,
  frame=none,
  escapeinside={(*@}{@*)}
}

\newtcblisting{promptbox}[1]{
  colback=gray!5!white,
  colframe=black!75!black,
  title={#1},
  fonttitle=\bfseries,
  sharp corners=south,
  boxrule=0.6pt,
  arc=1mm,
  left=2mm,
  right=2mm,
  top=1mm,
  bottom=1mm,
  width=\linewidth,
  breakable,
  listing only,
  listing options={style=promptstyle}
}

\title{\textsc{CultureBot}: A Framework for Retrieval-Augmented Exploration of Cultural Heritage Data}

\author{
    Vasileios Karampinis, Angelos Vlachos, Giorgos Filandrianos, \\
    \textbf{Maria Lymperaiou, Paraskevi Tzouveli, Chrysoula Zerva, Giorgos Stamou} \\
    School of Electrical and Computer Engineering, AILS Laboratory \\
    National Technical University of Athens \\
    \texttt{
        \{\href{mailto:vkarampinis@ails.ece.ntua.gr}{vkarampinis},
        \href{mailto:angelosv@ails.ece.ntua.gr}{angelosv},
        \href{mailto:geofila@ails.ece.ntua.gr}{geofila},
        \href{mailto:marialymp@ails.ece.ntua.gr}{marialymp}\}@ails.ece.ntua.gr
    } \\
    \texttt{
        \{\href{mailto:ptzouveli@ails.ece.ntua.gr}{ptzouveli},
        \href{mailto:czerva@ails.ece.ntua.gr}{czerva}\}@ails.ece.ntua.gr
    } \\
    \href{mailto:gstam@cs.ntua.gr}{\texttt{gstam@cs.ntua.gr}}
}

\begin{document}
\maketitle
\begin{abstract}
Digital cultural heritage collections encode rich relations among objects, places, periods, providers, images, and source institutions, yet users typically access them through keyword search and faceted browsing. We present \textsc{CULTUREBOT}, an interactive framework for natural-language exploration of cultural heritage collections that combines knowledge-graph-guided retrieval, lexical–semantic retrieval, and source-linked generation. \textsc{CULTUREBOT} maps user queries to structured catalogue relations, enriches retrieved records with textual evidence, and generates answers that link back to the underlying repository records. We instantiate the framework on 81,852 SearchCulture records and release an online demonstration. In an expert evaluation involving six cultural-heritage specialists, 55 persona-grounded queries, and 250 evaluations, the graph-guided system was rated significantly higher than two general-purpose assistants and a non-graph ablation, with the largest gains in evidence grounding, diversity, reasoning, and completeness. These results support graph-guided retrieval as a practical way to expose the relational structure of cultural collections through conversational access while preserving links to catalogue evidence.
\end{abstract}

\section{Introduction}
\label{sec:introduction}

Digital cultural heritage collections bring together objects, images, archival descriptions, archaeological records, and institutional metadata from many providers. They are not simple text corpora: they encode relations between cultural objects, places, periods, materials, concepts, rights, digital representations, and source institutions. Semantic models such as the Europeana Data Model (EDM) and CIDOC CRM were introduced precisely to support this kind of heterogeneous cultural representation across collections \citep{doerr2003cidoc,charles2013carare}. Yet, for most users, this semantic richness remains hidden by keyword search, filters, and faceted browsing. These interfaces limit what a query can express and they work well only for users who already know the underlying semantic structure, the right vocabulary, metadata fields, spelling conventions, and query strategy before exploring.

Users can rarely rely on single keywords to query heritage collections, they need to ask relational and interpretive questions about the collection's entities and their relations, seeking comparison and synthesis rather than a simple lookup function \cite{exploratory}. These questions are difficult to express through a search box, but they are also too sensitive to leave to unconstrained language generation. Cultural records carry evidential weight: dates may be approximate, attributions uncertain, places historically ambiguous, and descriptions shaped by institutional cataloguing practices. At the same time, general-purpose assistants may improve fluency, but they can miss fine-grained archaeological constraints and weaken the connection between answers and catalogue evidence. A fluent but unsupported answer can therefore create misleading historical connections, obscure the evidential basis of interpretation, and distance users from the original repositories.

\begin{figure*}[t]
    \centering
    \includegraphics[width=0.92\linewidth]{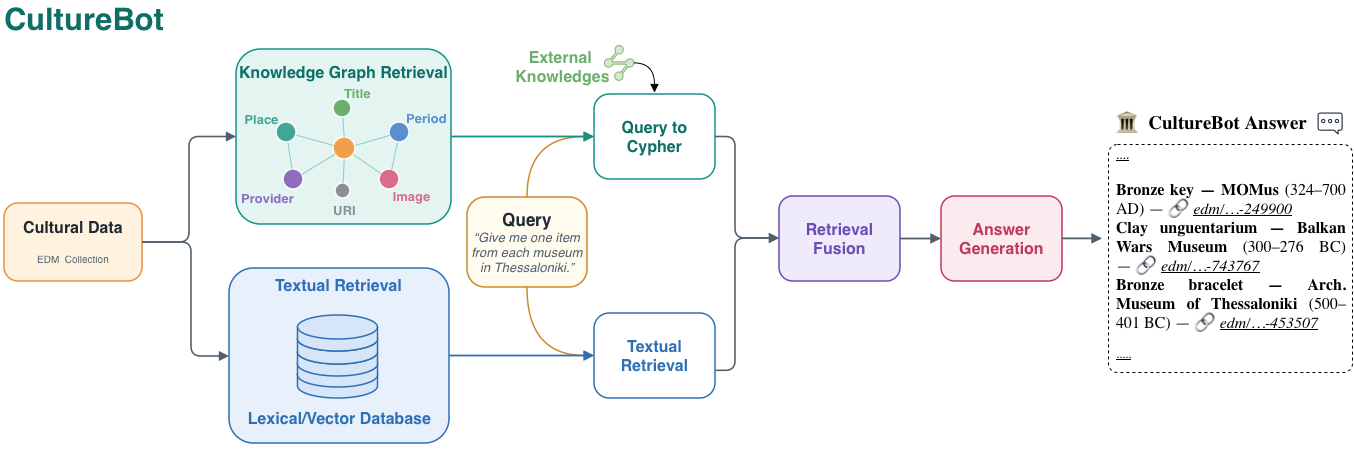}
    \caption{Overview of the \textsc{CultureBot} architecture.}
    \label{fig:architecture}
\end{figure*}

To address these concerns, we present \textsc{CultureBot}\footnote{\url{https://geofila.github.io/CultureBot/}}, a reusable framework for language-driven exploration of cultural heritage data. Rather than being tied to a single repository, \textsc{CultureBot} is designed for cultural datasets that combine structured metadata, textual descriptions, source records, rights information, images, and repository links. Given a user query, \textsc{CultureBot} supports structured search over cultural metadata, lexical--semantic retrieval over record descriptions, and source-linked generation over the retrieved records.

The contribution of \textsc{CultureBot} is threefold: 1) 
We introduce a reusable conversational framework for exploring heterogeneous cultural-heritage collections through natural language while retaining direct links to the underlying catalogue evidence; 2) We design a graph-centered retrieval pipeline that jointly exploits structured cultural metadata and lexical–semantic evidence, enabling queries involving relations, hierarchies, grouping, temporal constraints, and thematic descriptions that are difficult to express through conventional catalogue interfaces; 3) We release and demonstrate an interactive implementation instantiated on SearchCulture, called \textsc{CultureBot}\footnote{\href{https://culturebot.ails.ece.ntua.gr}{culturebot.ails.ece.ntua.gr}; a dedicated account for EACL evaluation is available without personal registration (email: \texttt{culturebotreviewer@ails.ece.ntua.gr}; password: \texttt{culturebotreviewer@eacl2026!}).} and validate its utility through 250 expert evaluations across realistic cultural-heritage information needs. This evaluation examines whether graph-grounded retrieval improves the usefulness, factuality, grounding, and role-specific adequacy of generated answers for cultural heritage exploration over both general-purpose assistants (ChatGPT, Gemini), and a text-only grounded variant of our approach (\textsc{CultureBot} w/o KG).

\section{Related Work}
\label{sec:related-work}

Retrieval-augmented generation (RAG) is a natural starting point for evidence-conditioned answering \citep{lewis2020rag,karpukhin2020dpr,gao2023ragsurvey}, and work on hallucination and factuality shows the need to verify generated claims against external evidence \citep{ji2023survey,asai2024selfrag,min2023factscore}. However, cultural heritage evidence is not only textual. It is distributed across descriptions, metadata fields, source records, object--place--period relations, rights statements, images, and institutional provenance. A standard retriever may find a relevant passage, but it does not necessarily preserve the structured distinctions that matter for historical interpretation.

Knowledge-graph question answering, text-to-query methods, and GraphRAG approaches show that relational structure can support question answering beyond flat document retrieval \citep{usbeck2024qald10,schneider2024semanticparsing,feng2024cypherbench,edge2024graphrag}. Recent cultural heritage systems make this direction more concrete: museum and heritage knowledge graphs integrate heterogeneous artifact metadata and support structured access \citep{carriero2019arco,li2025musekg,barzaghi2025chadkg,sartini2024iicongraph}; provenance-oriented RAG systems explore natural-language access to historically sensitive archival records \citep{henrickson2025gettyrag,scharnhorst2026cocreation}; and multimodal interfaces support visual search and discovery in digital collections \citep{huang2025digitalcollections,arnold2024visualheritage}. 

These works demonstrate the value of structure, retrieval, provenance, and visual exploration for cultural data.
However, they still leave open a synthesis problem: how to generate useful answers from heterogeneous cultural evidence while preserving the distinction between textual descriptions, metadata relations, source records, images, rights, and uncertain historical interpretation. \textsc{CultureBot} addresses this gap by using a cultural knowledge graph as the main access layer and by treating generation as source-linked synthesis over graph-organized cultural records.

\section{System Design}
\label{sec:methodology}

\subsection{Overview}

\textsc{CultureBot} is a knowledge-graph-centered conversational system for exploring cultural heritage collections. Its design reflects the fact that cultural records combine structured relations---such as objects, places, periods, providers, concepts, rights, and source institutions---with textual descriptions and digital representations.

Given a natural-language query, \textsc{CultureBot} first grounds relevant entities against the cultural knowledge graph and auxiliary vocabularies. It then retrieves records through structured metadata relations, enriches them with textual evidence, and generates a source-linked response over the fused retrieval context. Figure~\ref{fig:architecture} summarizes the pipeline.

\subsection{Knowledge-Graph Retrieval}

Let $\mathcal{C}$ be the collection. Each record $r \in \mathcal{C}$ consists of an identifier, structured metadata, textual fields, and a source URI, written compactly as $r=\langle id,m,t,u\rangle$. \textsc{CultureBot} projects the structured metadata into a graph $G=(V,E)$, where nodes represent cultural entities and edges represent typed metadata relations. Node types are derived automatically from the EDM class of each record (in jsonld format) and comprise \texttt{Object} (ProvidedCHO), \texttt{Place}, \texttt{TimeSpan}, \texttt{Agent}, \texttt{Concept} and \texttt{Aggregation}, together with \texttt{External} nodes that stand for entries in external authority files. Provider, rights, medium, image, landing-page, and source-identifier information is preserved as node properties on the \texttt{Object} and \texttt{Aggregation} nodes.

The graph preserves metadata roles that plain text retrieval tends to blur. For example, a place may be what an object depicts (its subject), where the object was found or produced, where its holding institution is located, or merely a word in a free-text description. To a text index all of these are the same string, in the graph they are distinct relations. This matters for queries that require hierarchy, grouping, or structured constraints, such as \emph{``give me one item from each museum in the Thessaloniki area''}. More complex queries may additionally restrict the results by period or object type, for example asking for items from the fifth century or from a specific category. A text-only retrieval system may find records mentioning ``Thessaloniki'', but it cannot reliably expand the area, group by museum, enforce one item per group, and apply such metadata constraints accurately.

\textsc{CultureBot} can further enrich the graph with external RDF resources, place hierarchies, authority files, and controlled vocabularies.\footnote{e.g. \href{https://www.semantics.gr/authorities/vocabularies}{https://www.semantics.gr/authorities/vocabularies}} During query processing, fuzzy matching over these resources identifies candidate entities relevant to the user query. These candidates are passed to the query prompt together with the graph schema, so that the text-to-query module operates over grounded cultural entities rather than guessing domain structure from the query alone.

\subsubsection{Knowledge-Graph Description}
\label{sec:kg-description}
The graph is built automatically from the structured EDM metadata; no label or edge type is defined manually. Every record with an identifier becomes a node carrying a base \texttt{Entity} label plus a type label taken directly from its EDM class, so the label
vocabulary follows the source data. Multilingual fields are flattened into node properties; for instance, free-text dates (e.g.\ ``101 BC -- 200 AD'') are parsed into integer year bounds that are indexed for date-range constraints (e.g. -101, 200).

Edges encode the metadata relations declared in the source. An Object links to its subjects (\texttt{HAS\_SUBJECT}), object categories (\texttt{HAS\_TYPE}), place (\texttt{LOCATED\_IN}), and period (\texttt{HAS\_TEMPORAL\_REFERENCE}); \texttt{Aggregation} nodes connect to the objects they describe (\texttt{AGGREGATES}); places nest via \texttt{IS\_PART\_OF}; and concepts form a SKOS hierarchy (\texttt{BROADER\_THAN}, \texttt{NARROWER\_THAN}, \texttt{BROAD\_MATCH}). Links to external authorities are typed edges to \texttt{External} nodes (\texttt{EXACT\_MATCH}, \texttt{SAME\_AS}), grounding internal concepts and places in resources such as GeoNames\footnote{e.g. \url{https://www.geonames.org/}}.

\subsubsection{Query Grounding and Graph Retrieval}

For each user query, \textsc{CultureBot} first builds a graph-query context instead of sending the query directly to the answer-generation model. This context includes the original query, a compact description of the graph schema, and candidate entities retrieved through fuzzy matching over internal labels and external vocabularies. For example, a mention such as ``Thessaloniki area'' may be expanded into candidate places, providers, or museums before graph querying.

\textsc{CultureBot} then uses an LLM to generate a read-only Cypher query from this grounded context. The LLM receives the user query, the schema, and the matched candidate entities, and is instructed to express the information need as graph operations over the available nodes and relations. This allows constraints such as place hierarchy, provider grouping, period filtering, object category, image availability, and source links to be handled through the graph rather than through text matching alone. 

Before execution, the generated Cypher query is validated. It must be syntactically valid, read-only, and compatible with the available graph schema; unsafe operations that create, delete, or modify graph content are rejected. The output of this stage is a set of graph-selected records that satisfy the structured constraints identified in the user query. Details of the validation process are provided in Appendix~\ref{app:query-validation}.

\subsection{Textual Retrieval}

Graph retrieval identifies candidate records, but generation also requires descriptive evidence. \textsc{CultureBot} therefore enriches graph-selected records using hybrid textual retrieval over record renderings. Lexical retrieval captures exact names, dates, identifiers, and domain terms, while dense retrieval captures paraphrases and broader semantic similarity. Rather than combining the two raw scores directly, which would require normalizing the incomparable scales of lexical and dense similarity, \textsc{CultureBot} fuses the two ranked lists through weighted reciprocal rank fusion (RRF),
\begin{equation}
\begin{aligned}
s(q,r) ={}& \alpha\,\frac{1}{k+\mathrm{rank}_{\mathrm{dense}}(q,r)} \\
       +{}& (1-\alpha)\,\frac{1}{k+\mathrm{rank}_{\mathrm{lex}}(q,r)},
\end{aligned}
\label{Eq:RRF}
\end{equation}
where $q$ is the user query, $r$ is a candidate record and $\mathrm{rank}_{\mathrm{lex}}(q,r)$ and $\mathrm{rank}_{\mathrm{dense}}(q,r)$ are the positions of $r$ in the lexical and dense ranked lists, respectively. The constant $k$ controls how quickly a record's contribution decays with its rank, while $\alpha \in [0,1]$ sets the relative weight of the two signals. A record that appears in both ranked lists retains both terms, whereas one retrieved by a single method receives only the corresponding term.

This component is not a separate alternative to the graph pipeline. It supplies the textual context needed to explain graph-selected records and improves recall when parts of the query are thematic or descriptive rather than directly expressible as graph constraints.

\subsection{Retrieval Fusion}

The graph and textual retrieval components return complementary information. The graph component identifies records that satisfy the structured constraints of the query, while textual retrieval adds the descriptive fields needed to explain those records. \textsc{CultureBot} combines these outputs by merging duplicate records, preserving source URIs, and attaching both metadata relations and textual snippets to each selected item.

When the query contains explicit structural constraints, such as place hierarchy, provider grouping, period, object type, or image availability, graph-selected records are treated as the primary result set. Textual retrieval is then used to enrich these records with titles, descriptions, notes, and other human-readable context. When the query is more thematic or descriptive, textual retrieval can also add relevant records that were not selected directly through graph constraints.

\subsection{Answer Generation}
\label{sec:answer_generation}

The final answer is generated from the user query and the fused retrieval context. The goal is not only to produce a fluent response, but to guide the user back to the original repository records. \textsc{CultureBot} is therefore instructed to answer from the selected records and to include source references inside the generated text, so that users can see which record supports each part of the answer. When available, these references include record titles, identifiers, and source URIs that lead back to the original repository entries.

The generation module is also constrained to avoid inventing dates, places, object types, providers, source links, or historical relations. When the indexed collection does not contain sufficient information to answer the query, the system is instructed to abstain rather than produce an unsupported answer. In this way, source linking is treated as part of the answer itself, not as a detached list of links appended after generation. The prompts used at each step, together with the parameters needed to reproduce
the experiments, are included with the released system.

\subsection{Cultural Heritage Dataset}
\label{sec:online-demo}

We provide an online \textsc{CultureBot} demo instantiated on SearchCulture data. The current implementation uses 81,852 records from SearchCulture\footnote{\href{https://www.searchculture.gr}{searchculture.gr}} related to the monuments of the Directorate of Mobile Monuments of the Hellenic Ministry of Culture. For this instance, \textsc{CultureBot} builds a graph database over the structured metadata and a hybrid textual index over record renderings. The graph component uses the schema, fuzzy-matched entities, and selected external knowledge links for query planning, while the generation component receives metadata relations, textual snippets, source URIs, and image links when available.

Although the demo is instantiated on SearchCulture, the pipeline is repository-independent. It can be applied to other EDM-style cultural collections, or to cultural datasets that can be mapped to the \textsc{CultureBot} EDM-derived graph schema of Section~\ref{sec:kg-description}.

\section{Evaluation Setup}
\label{sec:evaluation-setup}



We evaluate \textsc{CultureBot} through an expert-based study addressing two research questions: (i) how its collection-grounded answers compare with those of general-purpose assistants, and (ii) to what extent graph-guided retrieval improves answer quality beyond textual retrieval alone. Six domain experts, including archaeologists and cultural heritage researchers familiar with SearchCulture and the evaluated collections, participated in the study. They developed 12 personas representing diverse users of cultural heritage collections, including researchers, students, educators, curators, tour guides, and artists. Based on these personas, the experts authored 55 queries spanning factual lookup, spatial exploration, thematic comparison, object selection, image-oriented requests, and multi-step reasoning over cultural records.

For each query, we compared four systems: ChatGPT, Gemini, \textsc{CultureBot} without the knowledge graph (\textsc{CultureBot}~w/o KG), and the full graph-guided system (\textsc{CultureBot}~w/ KG). ChatGPT and Gemini served as general-purpose baselines, reflecting tools that the participating experts reported using in practice. The comparison between the two \textsc{CultureBot} variants isolates the contribution of graph-guided retrieval, since they share the same textual retrieval and answer-generation components. To ensure that judgments focused on answer quality rather than interface differences, outputs from all four systems were presented through the same annotation platform. The public \textsc{CultureBot} demo uses Open WebUI, an established interface previously evaluated in user studies \cite{baek2025open}.

\paragraph{Annotation protocol.}
The resulting answers were evaluated through a tailor-made annotation platform. Each expert was presented with the original persona, the query, and the four system answers. In total, the study comprises 250 evaluation sessions over 55  authored queries. The platform also included a query-level assessment based on manual use of SearchCulture. Experts estimated how difficult each query would be to answer through simple SearchCulture search, how much time such a manual search would take, and which SearchCulture query or filters they used when attempting the search.

\paragraph{Evaluation criteria.}
Answer quality was assessed using a twelve-criterion rubric covering general dimensions (query relevance, clarity, result usefulness, justification, hallucination, and bias) alongside dimensions particularly relevant to cultural heritage exploration (evidence grounding, diversity, completeness, required reasoning steps, usefulness relative to manual SearchCulture search, and need for further search). All criteria were mapped to a common $0$--$3$ scale, with higher scores indicating better quality and the need for further search reverse-coded. The complete rubric and response options are provided in Appendix~\ref{app:expert-participants}.

\section{Evaluation results}
\label{sec:evaluation}

Averaged over all criteria (Table \ref{tab:overall-quality}), \textsc{CultureBot} with knowledge graph (\textsc{CultureBot} w/ KG) received the highest quality score (2.34 out of 3), followed by ChatGPT (2.17), \textsc{CultureBot} without knowledge graph (\textsc{Culturebot} w/o KG) (2.00), and Gemini (1.82). Since the systems were rated on the same queries, we also compared them pairwise within each query: \textsc{CultureBot} w/ KG was rated significantly above all three alternatives, ChatGPT ($p = 0.001$), \textsc{CultureBot} w/o KG ($p < 0.001$), and, by the widest margin, Gemini ($p < 0.001$; all $p$-values Holm-corrected over the $n = 110$ evaluations in which all four systems were rated).

\begin{table}[!ht]
\centering
\small
\caption{Overall answer quality (mean $\pm$ SD; each score averages twelve criteria; $n$ = rated answers). Holm-corrected Wilcoxon signed-rank $p$-values compare each baseline with \textsc{CultureBot} w/ KG \citep{wilcoxon1945}; best values are in \textbf{bold}.}

\label{tab:overall-quality}
\resizebox{\columnwidth}{!}{%
\begin{tabular}{lccc}
\toprule
System & Mean quality & $\#n$ & $p$ \\
\midrule
ChatGPT & 2.17 {\scriptsize $\pm\,0.39$} & 112 & $p = 0.001$ \\
Gemini & 1.82 {\scriptsize $\pm\,0.67$} & 112 & $p < 0.001$ \\ \midrule
\textsc{CultureBot} w/o KG & 2.00 {\scriptsize $\pm\,0.72$} & 116 & $p < 0.001$ \\
\textsc{CultureBot} w/ KG & \textbf{2.34} {\scriptsize $\pm\,0.55$} & 110 & --- \\
\bottomrule
\end{tabular}}
\end{table}

Looking at individual criteria (Table \ref{tab:per-criterion}), the systems differ in substance far more than in style. All criteria except bias significantly separate the systems (Friedman test, p < 0.05). Separation is concentrated in evidence grounding, the largest difference of all, with \textsc{CultureBot} w/ KG at 2.59 versus Gemini at 1.46, followed by diversity, reasoning, and completeness. In contrast, clarity and result usefulness separate the systems only weakly and bias not at all: all four produce fluent, largely unbiased text. Notably, Gemini matches the best hallucination systems (2.90) while being by far the least grounded: it avoids stating anything wrong mainly by committing to little, producing safe but thin answers. Conversely, the non-graph variant (\textsc{CultureBot} w/o KG) is the only system with a substantial hallucination rate 23\% of its determinable ratings were flagged as minor or major hallucination, against 6\% for every other system, including \textsc{CultureBot} w/ KG, which shares its textual index and generation model. 

\begin{table}[!ht]

\centering

\small

\caption{Mean score (0--3) per rubric criterion and system. Best values are shown in \textbf{bold}.}

\label{tab:per-criterion}

\resizebox{\columnwidth}{!}{%

\begin{tabular}{l|cc|cc}

\toprule

Criterion & ChatGPT & Gemini & \textsc{CB} w/o KG & \textsc{CB} w/ KG \\

\midrule

Evidence grounding & 2.05 & 1.46 & 2.09 & \textbf{2.59} \\

Diversity & 2.17 & 1.68 & 2.06 & \textbf{2.42} \\

Reasoning & 2.36 & 1.79 & 2.09 & \textbf{2.53} \\

Further search (rev.) & 1.46 & 1.21 & 1.51 & \textbf{1.85} \\

Completeness & 1.84 & 1.46 & 1.66 & \textbf{2.18} \\

Useful.\ vs manual & 1.75 & 1.40 & 1.59 & \textbf{1.92} \\

Query relevance & \textbf{2.46} & 2.12 & 2.03 & 2.38 \\

Hallucination & \textbf{2.91} & 2.90 & 2.53 & 2.90 \\

Justification & 2.04 & 1.68 & 1.84 & \textbf{2.15} \\

Result usefulness & 2.04 & 1.67 & 1.81 & \textbf{2.12} \\

Clarity & 2.25 & 1.91 & 2.03 & \textbf{2.29} \\

\midrule

\textit{Bias (absence)} & 2.79 & 2.66 & 2.74 & \textbf{2.85} \\

\bottomrule

\end{tabular}}

\end{table}

Experts also rated each query's difficulty for manual SearchCulture search (median 3 of 5) and its estimated search time (median 5 minutes, interquartile range 3-15) against near-immediate system responses (Section~\ref{sec:evaluation-setup}). Stratified by this rating (Table~\ref{tab:difficulty}), \textsc{CultureBot}'s advantage concentrates on easy and moderate queries (2.43 vs.\ 2.08 for ChatGPT), while on the hardest the systems converge (2.26 vs.\ 2.22): such queries map poorly onto the catalogue metadata, leaving the graph little leverage.

\begin{table}[!ht]
\centering
\small
\caption{Answer quality (0--3) by expert-estimated difficulty of manual SearchCulture search.}
\label{tab:difficulty}
\resizebox{\columnwidth}{!}{%
\begin{tabular}{l|cc|cc}
\toprule
Difficulty & ChatGPT & Gemini & \textsc{CB} w/o KG & \textsc{CB} w/ KG \\
\midrule
Easy (1--2)  & 2.08 & 1.63 & 1.95 & \textbf{2.43} \\
Moderate (3) & 2.23 & 1.77 & 1.95 & \textbf{2.37} \\
Hard (4--5)  & 2.22 & 1.97 & 2.06 & \textbf{2.26} \\
\bottomrule
\end{tabular}}
\end{table}

The experts overall attitude points the same way (Table~\ref{tab:attitude}). \textsc{CultureBot} w/ KG was the only system rated significantly above parity with manual SearchCulture search with ChatGPT (3.33) and \textsc{CultureBot} w/o KG (3.12) near parity and Gemini alone below it (2.86). \textsc{CultureBot} w/ KG answers were also judged fully sufficient in about a quarter of cases, compared with 4\% for Gemini and 9\% for ChatGPT, whose answers most often sent experts back to the platform for additional examples and images (62\% of cases). Still, some follow-up search usually remained desirable: fully displacing manual search is an open goal. Attitude was shaped by completeness, usefulness, and diversity far more than by the absence of hallucination or bias, experts treat factual accuracy as a prerequisite and reward
coverage.

\begin{table}[!ht]
\centering
\small
\caption{Expert attitude per system. Best values are shown in \textbf{bold}.}
\label{tab:attitude}
\resizebox{\columnwidth}{!}{%
\begin{tabular}{lccc}
\toprule
System & Usefulness & SD & ``Answer sufficient'' \\
\midrule
ChatGPT & 3.33 & 1.29 & 9.1\% \\
Gemini & 2.86 & 1.47 & 3.6\% \\ \midrule
\textsc{CultureBot} w/o KG & 3.12 & 1.35 & 16.4\% \\
\textsc{CultureBot} w/ KG & \textbf{3.55} & 1.11 & \textbf{24.5\%} \\
\bottomrule
\end{tabular}}
\end{table}

%


\section{Conclusion}
We presented \textsc{CultureBot}, a reusable framework for language-driven exploration of cultural heritage collections that keeps generated answers linked to the original catalogue records. An expert evaluation on SearchCulture data (250 evaluations, six experts) supports three conclusions. First, graph-grounded retrieval measurably improves answer quality: the full system was rated significantly above both general-purpose assistants and its non-graph ablation, so the gain comes from graph-guided access, not from generation alone. Second, the improvement lies in substance, not style: the systems differ most in grounding, diversity, reasoning, and completeness, and barely in clarity or bias; one assistant paired top factuality with the weakest grounding, giving safe but thin answers. Third, expert attitude follows evidence: the graph-guided system was rated the most useful against manual search, sufficient in about a quarter of cases, with attitude tracking completeness and diversity rather than the absence of hallucination. More broadly, the relational structure hidden behind keyword search is precisely what a conversational assistant needs to ground its answers.

\paragraph{Limitations.}
Our evaluation is limited to a single cultural heritage collection and expert judgments of generated answers, rather than end-to-end system usability or downstream user outcomes. Although the evaluation spans 12 personas, the limited number of queries per role prevents robust conclusions about the needs of individual user groups. Different users may also require capabilities beyond the current system, such as visual search for artists and designers, audience-adapted explanations for educators, and more detailed provenance information for curators and heritage officers. The reduced advantage on the most difficult queries further highlights the system's dependence on the coverage and quality of catalogue metadata. Moreover, all evaluation queries were in English, despite the underlying records being predominantly in Greek, leaving multilingual performance untested. Future work should assess transfer to other collections, evaluate the live system in realistic user workflows, and investigate its impact on learning and professional practice.


\bibliography{custom}

@article{exploratory,
author = {Marchionini, Gary},
title = {Exploratory search: from finding to understanding},
year = {2006},
issue_date = {April 2006},
publisher = {Association for Computing Machinery},
address = {New York, NY, USA},
volume = {49},
number = {4},
issn = {0001-0782},
url = {https://doi.org/10.1145/1121949.1121979},
doi = {10.1145/1121949.1121979},
journal = {Commun. ACM},
month = apr,
pages = {41–46},
numpages = {6}
}

@article{doerr2003cidoc,
author = {Doerr, Martin},
title = {The CIDOC conceptual reference module: an ontological approach to semantic interoperability of metadata},
year = {2003},
issue_date = {Fall 2003},
publisher = {John Wiley \& Sons, Inc.},
address = {USA},
volume = {24},
number = {3},
issn = {0738-4602},
journal = {AI Mag.},
month = sep,
pages = {75–92},
numpages = {18}
}

@misc{charles2013carare,
      title={Achieving interoperability between the CARARE schema for monuments and sites and the Europeana Data Model}, 
      author={Valentine Charles and Antoine Isaac and Kate Fernie and Costis Dallas and Dimitris Gavrilis and Stavros Angelis},
      year={2013},
      eprint={1306.2869},
      archivePrefix={arXiv},
      primaryClass={cs.DL},
      url={https://arxiv.org/abs/1306.2869}, 
}

@inproceedings{lewis2020rag,
author = {Lewis, Patrick and Perez, Ethan and Piktus, Aleksandra and Petroni, Fabio and Karpukhin, Vladimir and Goyal, Naman and K\"{u}ttler, Heinrich and Lewis, Mike and Yih, Wen-tau and Rockt\"{a}schel, Tim and Riedel, Sebastian and Kiela, Douwe},
title = {Retrieval-augmented generation for knowledge-intensive NLP tasks},
year = {2020},
isbn = {9781713829546},
publisher = {Curran Associates Inc.},
address = {Red Hook, NY, USA},
booktitle = {Proceedings of the 34th International Conference on Neural Information Processing Systems},
articleno = {793},
numpages = {16},
location = {Vancouver, BC, Canada},
series = {NIPS '20}
}

@inproceedings{karpukhin2020dpr,
    title = "Dense Passage Retrieval for Open-Domain Question Answering",
    author = "Karpukhin, Vladimir  and
      Oguz, Barlas  and
      Min, Sewon  and
      Lewis, Patrick  and
      Wu, Ledell  and
      Edunov, Sergey  and
      Chen, Danqi  and
      Yih, Wen-tau",
    editor = "Webber, Bonnie  and
      Cohn, Trevor  and
      He, Yulan  and
      Liu, Yang",
    booktitle = "Proceedings of the 2020 Conference on Empirical Methods in Natural Language Processing (EMNLP)",
    month = nov,
    year = "2020",
    address = "Online",
    publisher = "Association for Computational Linguistics",
    url = "https://aclanthology.org/2020.emnlp-main.550/",
    doi = "10.18653/v1/2020.emnlp-main.550",
    pages = "6769--6781"
}

@article{gao2023ragsurvey,
author = {Huang, Yizheng and Huang, Jimmy Xiangji},
title = {A Survey on Retrieval-Augmented Text Generation for Large Language Models},
year = {2026},
issue_date = {September 2026},
publisher = {Association for Computing Machinery},
address = {New York, NY, USA},
volume = {58},
number = {12},
issn = {0360-0300},
url = {https://doi.org/10.1145/3805774},
doi = {10.1145/3805774},
journal = {ACM Comput. Surv.},
month = may,
articleno = {300},
numpages = {38}
}

@article{ji2023survey,
author = {Ji, Ziwei and Lee, Nayeon and Frieske, Rita and Yu, Tiezheng and Su, Dan and Xu, Yan and Ishii, Etsuko and Bang, Ye Jin and Madotto, Andrea and Fung, Pascale},
title = {Survey of Hallucination in Natural Language Generation},
year = {2023},
issue_date = {December 2023},
publisher = {Association for Computing Machinery},
address = {New York, NY, USA},
volume = {55},
number = {12},
issn = {0360-0300},
url = {https://doi.org/10.1145/3571730},
doi = {10.1145/3571730},
journal = {ACM Comput. Surv.},
month = mar,
articleno = {248},
numpages = {38}
}

@inproceedings{asai2024selfrag,
  title={Self-rag: Learning to retrieve, generate, and critique through self-reflection},
  author={Asai, Akari and Wu, Zeqiu and Wang, Yizhong and Sil, Avi and Hajishirzi, Hannaneh},
  booktitle={International conference on learning representations},
  volume={2024},
  pages={9112--9141},
  year={2024}
}

@inproceedings{min2023factscore,
  title={Factscore: Fine-grained atomic evaluation of factual precision in long form text generation},
  author={Min, Sewon and Krishna, Kalpesh and Lyu, Xinxi and Lewis, Mike and Yih, Wen-tau and Koh, Pang and Iyyer, Mohit and Zettlemoyer, Luke and Hajishirzi, Hannaneh},
  booktitle={Proceedings of the 2023 Conference on Empirical Methods in Natural Language Processing},
  pages={12076--12100},
  year={2023}
}

@article{edge2024graphrag,
  title={From local to global: A graph rag approach to query-focused summarization, 2025},
  author={Edge, Darren and Trinh, Ha and Newman Cheng, Joshua Bradley and Chao, Alex and Mody, Apurva and Truitt, Steven and Metropolitansky, Dasha and Ness, Robert Osazuwa and Larson, Jonathan},
  journal={URL https://arxiv. org/abs/2404.16130}
}

@article{usbeck2024qald10,
  title={QALD-10--The 10th challenge on question answering over linked data: Shifting from DBpedia to Wikidata as a KG for KGQA},
  author={Usbeck, Ricardo and Yan, Xi and Perevalov, Aleksandr and Jiang, Longquan and Schulz, Julius and Kraft, Angelie and M{\"o}ller, Cedric and Huang, Junbo and Reineke, Jan and Ngonga Ngomo, Axel-Cyrille and others},
  journal={Semantic Web},
  volume={15},
  number={6},
  pages={2193--2207},
  year={2024},
  publisher={SAGE Publications Sage UK: London, England}
}

@article{schneider2024semanticparsing,
  title={Evaluating large language models in semantic parsing for conversational question answering over knowledge graphs},
  author={Schneider, Phillip and Klettner, Manuel and Jokinen, Kristiina and Simperl, Elena and Matthes, Florian},
  journal={arXiv preprint arXiv:2401.01711},
  year={2024}
}

@inproceedings{feng2024cypherbench,
  title={Cypherbench: Towards precise retrieval over full-scale modern knowledge graphs in the llm era},
  author={Feng, Yanlin and Papicchio, Simone and Rahman, Sajjadur},
  booktitle={Proceedings of the 63rd Annual Meeting of the Association for Computational Linguistics (Volume 1: Long Papers)},
  pages={8934--8958},
  year={2025}
}

@inproceedings{carriero2019arco,
  title={ArCo: The Italian cultural heritage knowledge graph},
  author={Carriero, Valentina Anita and Gangemi, Aldo and Mancinelli, Maria Letizia and Marinucci, Ludovica and Nuzzolese, Andrea Giovanni and Presutti, Valentina and Veninata, Chiara},
  booktitle={International semantic web conference},
  pages={36--52},
  year={2019},
  organization={Springer}
}

@article{li2025musekg,
  title={MUSEKG: A Knowledge Graph Over Museum Collections},
  author={Li, Jinhao and Qi, Jianzhong and Han, Soyeon Caren and Holden, Eun-Jung},
  journal={arXiv preprint arXiv:2511.16014},
  year={2025}
}

@article{henrickson2025gettyrag,
  title={Retrieval-augmented generation for natural language art provenance searches in the Getty Provenance Index},
  author={Henrickson, Mathew and Atwell, Eric and Stell, John and Westgarth, Mark and Roy, Dibyadyuti and Abbas, Noorhan},
  journal={Academia AI and Applications},
  volume={2},
  number={1},
  year={2026},
  publisher={Academia. edu Journals}
}

@article{huang2025digitalcollections,
  title={Digital collections explorer: An open-source, multimodal viewer for searching digital collections},
  author={Huang, Ying-Hsiang and Lee, Benjamin Charles Germain},
  journal={Computational Humanities Research},
  volume={1},
  pages={e14},
  year={2025},
  publisher={Cambridge University Press}
}

@article{barzaghi2025chadkg,
  title={CHAD-KG: a knowledge graph for representing cultural heritage objects and digitisation paradata},
  author={Barzaghi, Sebastian and Moretti, Arianna and Heibi, Ivan and Peroni, Silvio},
  journal={arXiv preprint arXiv:2505.13276},
  year={2025}
}

@inproceedings{sartini2024iicongraph,
  title={IICONGRAPH: Improved iconographic and iconological statements in knowledge graphs},
  author={Sartini, Bruno},
  booktitle={European Semantic Web Conference},
  pages={57--74},
  year={2024},
  organization={Springer}
}

@article{arnold2024visualheritage,
  title={Explainable Search and Discovery of Visual Cultural Heritage Collections with Multimodal Large Language Models},
  author={Arnold, Taylor and Tilton, Lauren},
  year={2024}
}

@article{scharnhorst2026cocreation,
  title={Co-creation of AI technology, empowering curators of cultural heritage information and guarding research commons},
  author={Scharnhorst, Andrea and Yang, Han and Touber, Jetze and Ferguson, Kim and Mayr, Philipp and Tykhonov, Vyacheslav},
  journal={arXiv preprint arXiv:2605.28481},
  year={2026}
}

@incollection{wilcoxon1945,
  title={Individual comparisons by ranking methods},
  author={Wilcoxon, Frank},
  booktitle={Breakthroughs in statistics: Methodology and distribution},
  pages={196--202},
  year={1992},
  publisher={Springer}
}

@article{baek2025open,
  title={Open webui: An open, extensible, and usable interface for ai interaction},
  author={Baek, Jaeryang and Hussain, Ayana and Liu, Danny and Vincent, Nicholas and Kim, Lawrence H},
  journal={arXiv preprint arXiv:2510.02546},
  year={2025}
}

\clearpage

\appendix
\section{Query Validation}
\label{app:query-validation}

CultureBot uses an LLM to translate the grounded user query into a Neo4j Cypher query. Because this query is generated automatically, the implementation applies a validation step before execution. The goal of validation is to ensure that generated graph queries are read-only, syntactically well formed, and compatible with the graph-access pattern expected by the system. Further implementation details on the EDM-to-graph transformation, the resulting Neo4j schema, and its integration into \textsc{CultureBot} are provided in the released documentation.\footnote{\url{https://github.com/geofila/CultureBot/blob/main/knowledge_graph/README.md}}

The first validation step normalizes the generated output by removing possible Markdown code fences. The query is then required to start with \texttt{MATCH} or \texttt{OPTIONAL MATCH} and to contain a \texttt{RETURN} clause. Queries with unbalanced parentheses are rejected. The validator also blocks write, administrative, or bulk-loading operations, including \texttt{CREATE}, \texttt{MERGE}, \texttt{DELETE}, \texttt{SET}, \texttt{REMOVE}, \texttt{DROP}, \texttt{LOAD CSV}, \texttt{CALL dbms}, and write-oriented APOC procedures. These checks prevent the generated query from modifying the graph or invoking administrative functionality.

After textual validation, the query is checked against Neo4j using \texttt{EXPLAIN}. This step asks Neo4j to parse and plan the query without executing it. It therefore catches syntax errors and schema-level problems before any data are accessed. Only queries that pass both the read-only validation rules and the Neo4j \texttt{EXPLAIN} check are executed.

The Cypher-generation prompt also constrains the form of generated graph queries. It instructs the model to output only Cypher, to use \texttt{DISTINCT} when returning items, to include \texttt{LIMIT} for non-aggregate item queries, and to return item identifiers and labels when item records are requested. It also includes graph-schema information and few-shot examples verified against the live Neo4j schema. In particular, relationship-based filters are expressed through \texttt{EXISTS \{...\}} subqueries rather than through filtering conditions attached to \texttt{OPTIONAL MATCH}, since the latter can preserve rows with null optional variables and therefore fail to exclude unrelated records. This constraint is important for queries involving places, periods, materials, types, or other metadata relations.

If validation fails, the generated query is not executed and the system reports that the graph query could not be used. This conservative design prevents unsafe or malformed graph access and ensures that generated answers are based only on retrieval results produced by validated read-only queries.

\section{Implementation Details}
\label{app:implementation-details}

CultureBot is implemented as an OpenWebUI pipeline. The pipeline exposes configurable runtime parameters through OpenWebUI valves, allowing the same system to run in different retrieval and model configurations without changing the application code. The main generation model is selected through the \texttt{OPENAI\_MODEL} valve, while the endpoint can be configured through \texttt{OPENAI\_BASE\_URL}. This allows the implementation to use OpenAI-compatible model endpoints, including hosted models, Azure-style endpoints, or local proxy deployments. The current model used for the reported experiments and in the demo page at the time of the submission is: GPT-5.5\footnote{gpt-5.5-2026-04-23}.

The system supports three retrieval modes. In \texttt{kg} mode, CultureBot generates a read-only Cypher query, executes it against the Neo4j knowledge graph, and retrieves the corresponding catalogue records. In \texttt{rag} mode, it bypasses the graph and performs hybrid textual retrieval over record renderings. In \texttt{hybrid} mode, which is the default configuration, both graph retrieval and textual retrieval are used, and the retrieved context is merged into a single answer-generation prompt. This mode is used for the full CultureBot system because it combines structured graph constraints with descriptive textual evidence.

The textual retrieval component uses a hybrid FAISS and BM25 index. FAISS provides semantic retrieval over record renderings, while BM25 provides lexical matching for exact terms, names, dates, and identifiers. The two result lists are deduplicated and combined using reciprocal rank fusion. The implementation supports several embedding backends, including models from Hugging Face along with proprietary source models. These alternatives allow the system to trade off speed, local execution, multilingual quality, and API-based embedding quality. The embedding model used in the reported experiments is: \texttt{text-embedding-3-large}.

At answer-generation time, CultureBot assembles a bounded context from graph rows, matched catalogue records, textual snippets, source URIs, and image links when available. The number of graph rows, retrieved records, textual matches, and context length are controlled by runtime parameters such as \texttt{MAX\_GRAPH\_ROWS}, \texttt{MAX\_CONTEXT\_RECORDS}, \texttt{TOP\_K\_SEMANTIC}, \texttt{TOP\_K\_BM25}, and \texttt{MAX\_CONTEXT\_CHARS}. These limits are used to keep the prompt within the model context window while preserving enough retrieved material for multi-record answers.

\subsection{Demo Interface}

\begin{figure}[t]
    \centering
    \includegraphics[width=0.5\textwidth]{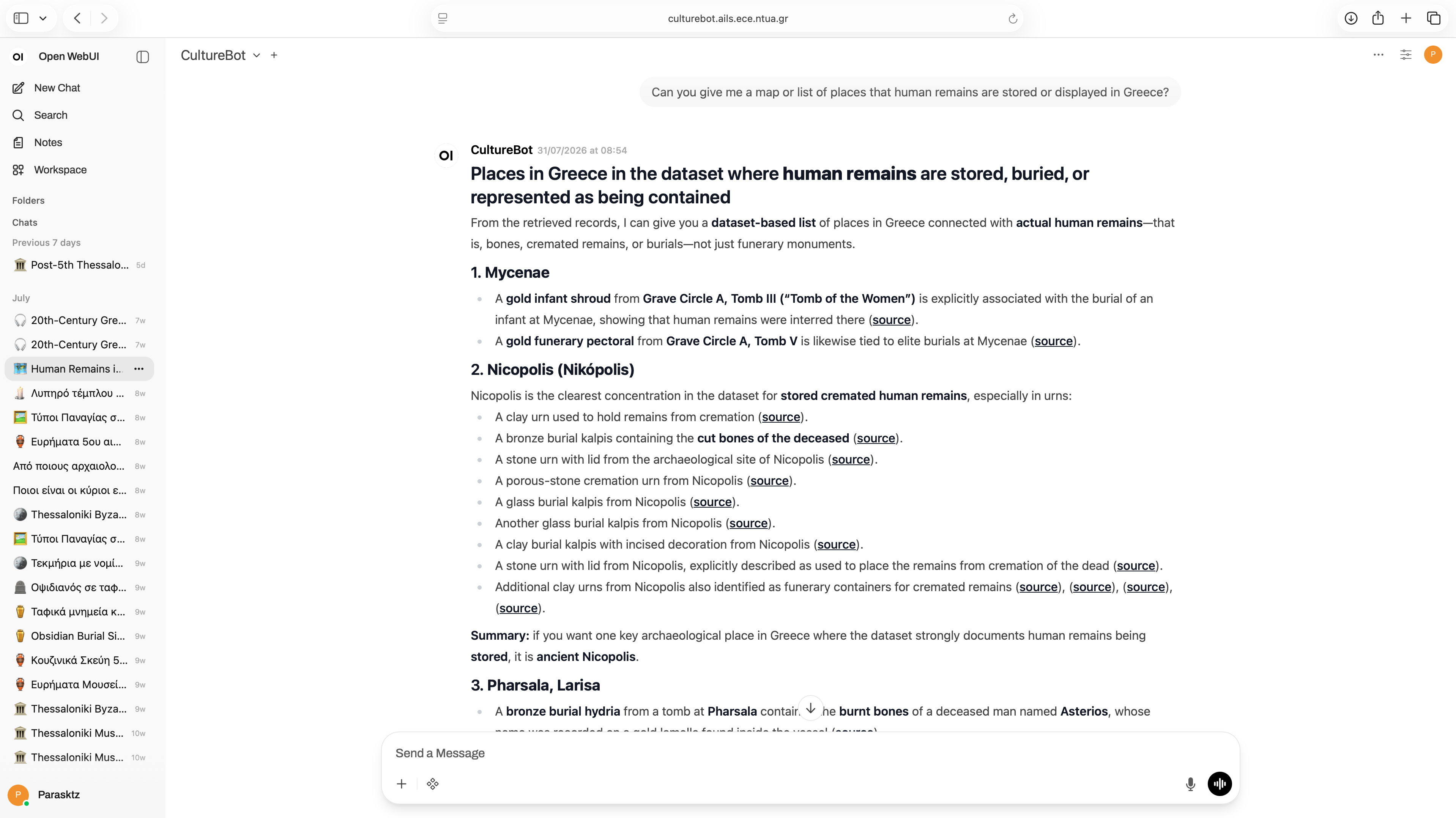}
    \caption{The \textsc{CultureBot} demo showing a query, its answer, and links to supporting catalogue records.}
    \label{fig:demo-interface}
\end{figure}

Figure~\ref{fig:demo-interface} shows the deployed \textsc{CultureBot} interface. Built on Open WebUI, the demo provides registration and sign-in and lets users explore cultural heritage records through a conversational interface. Users can revisit earlier conversations and continue them with the previous context retained. When supporting records are available, answers include links to the corresponding SearchCulture entries, allowing users to inspect the catalogue evidence behind the response. This makes the source-linked generation described in Section~\ref{sec:answer_generation} visible during use.

\section{Expert Participants and Evaluation Design}
\label{app:expert-participants}

The evaluation was designed with the support of six domain experts with substantial experience in cultural heritage documentation, archaeological interpretation, and the use of SearchCulture. The expert group included archaeologists, cultural heritage researchers, and collection specialists familiar with both the platform and the evaluated collections. Their role was twofold: first, to define realistic user scenarios and evaluation queries, and second, to assess the system outputs through a structured annotation protocol.

The evaluation set was organized around 12 expert-defined personas. These personas represent different groups that may use cultural heritage search systems, including researchers, students, educators, curators, tour guides, journalists, artists, craft professionals, and heritage officers. The experts authored 55 queries for these personas, covering factual, spatial, visual, comparative, thematic, and multi-step information needs.

For each query, answers were collected from four systems: ChatGPT\footnote{\texttt{gpt-5.6-sol}}, Gemini\footnote{\texttt{gemini-3.6-flash}}, CultureBot without the knowledge-graph component, and the full graph-guided CultureBot. ChatGPT and Gemini were selected because they were the general-purpose systems that the participating experts reported using in practice. The non-graph CultureBot variant was included to examine the contribution of the knowledge graph to retrieval and answer synthesis. Each answer was evaluated by at least 2 experts. 

\section{Personas}
\label{app:personas}

The evaluation was structured around 12 expert-defined personas representing distinct cultural heritage user groups and information needs. Each persona specifies a professional or educational role, a use context, and a concrete information objective, allowing the evaluation to capture variation in the evidence, reasoning, and output required by different users. Curators and educators seek representative and audience-appropriate material; researchers require precise metadata and relational evidence; artists and craft practitioners emphasize visual and formal properties; tour guides rely on spatial and itinerary-oriented information; and ethics officers examine provenance and culturally sensitive contexts. Grounding the evaluation in these scenarios moves it beyond generic question answering and tests whether the system can support heterogeneous forms of cultural heritage exploration. The resulting 55 queries span factual retrieval, spatial and visual search, comparison, synthesis, and multi-step reasoning. Full persona descriptions, use cases, and associated queries are available in the released evaluation materials.\footnote{\url{https://github.com/geofila/CultureBot/tree/main/evaluation}}

\end{document}